\documentclass[journal]{IEEEtran}

\usepackage{amsmath}
\usepackage{array}
\usepackage[utf8]{inputenc}
\usepackage[margin=1in]{geometry}
\usepackage{subcaption}
\usepackage{xcolor}
\usepackage{graphicx}
\usepackage{comment}
\usepackage{gensymb}
\usepackage{verbatim}
\usepackage{url}

\begin{document}


\title{Measuring a Robot Hand's Graspable Region using Power and Precision Grasps}
\author{John~Morrow*, Joshua Campbell*,  Nuha~Nishat,  Ravi~Balasubramanian, and~Cindy~Grimm \thanks{morrowjo, campbjos, nishatn, balasubr, grimmc at Oregon State University, *authors contributed equally} }

\maketitle


\begin{abstract}
The variety of robotic hand designs and actuation schemes makes it difficult to measure a hand's graspable volume. For end-users, this lack of standardized measurements makes it challenging to determine {\em a priori} if a robot hand is the right size for grasping an object. We propose a practical hand measurement standard, based on precision and power grasps, that is applicable to a wide variety of robot hand designs. The resulting measurements can be used to both determine if an object will fit in the hand {\em and} characterize the size of an object with respect to the hand. Our measurement procedure uses a {\em functional} approach, based on grasping a hypothetical cylinder, that allows the measurer choose the exact hand orientation and finger configurations that are used for the measurements. This ensures that the measurements are {\em functionally} comparable while relying on the human to determine the finger configurations that best match the intended grasp. We demonstrate using our measurement standard with three commercial robot hand designs and objects from the YCB data set.


\end{abstract}

\section{Introduction}
Fingered robot hands vary widely both in kinematic design and actuation. This diversity in morphology makes it difficult to compare the capabilities of various hands and choose the correct one for a specific grasping task. For example, a robot hand big enough to grasp a cantaloupe between a pair of fingers may not be able to close tight enough hold a hammer in a power grasp. 

Most robot hands come with design specs such as finger length and joint angle ranges, but these alone are insufficient because the actuation scheme means not all finger configurations are reachable. 
An alternative approach is evaluating the graspable space of the robot hand using dynamic simulation or physical experiments, but this is cumbersome and does not scale well. In this paper we propose a middle ground between these two approaches, in which someone familiar with the hand takes a small number of ``functional'' measurements for a specific grasp type (precision or power). 
An end user may use these measurements to determine if a specific object will fit anywhere within a hand's graspable space by linearly interpolating the measurements, providing an approximation of a hand's graspable space as it actuates. 
End users may also compute the object's size relative to the graspable space of the hand to assess how well the object fits inside the hand (small, medium, large, etc).

\begin{figure}
    \centering
    \includegraphics[width=0.95\linewidth]{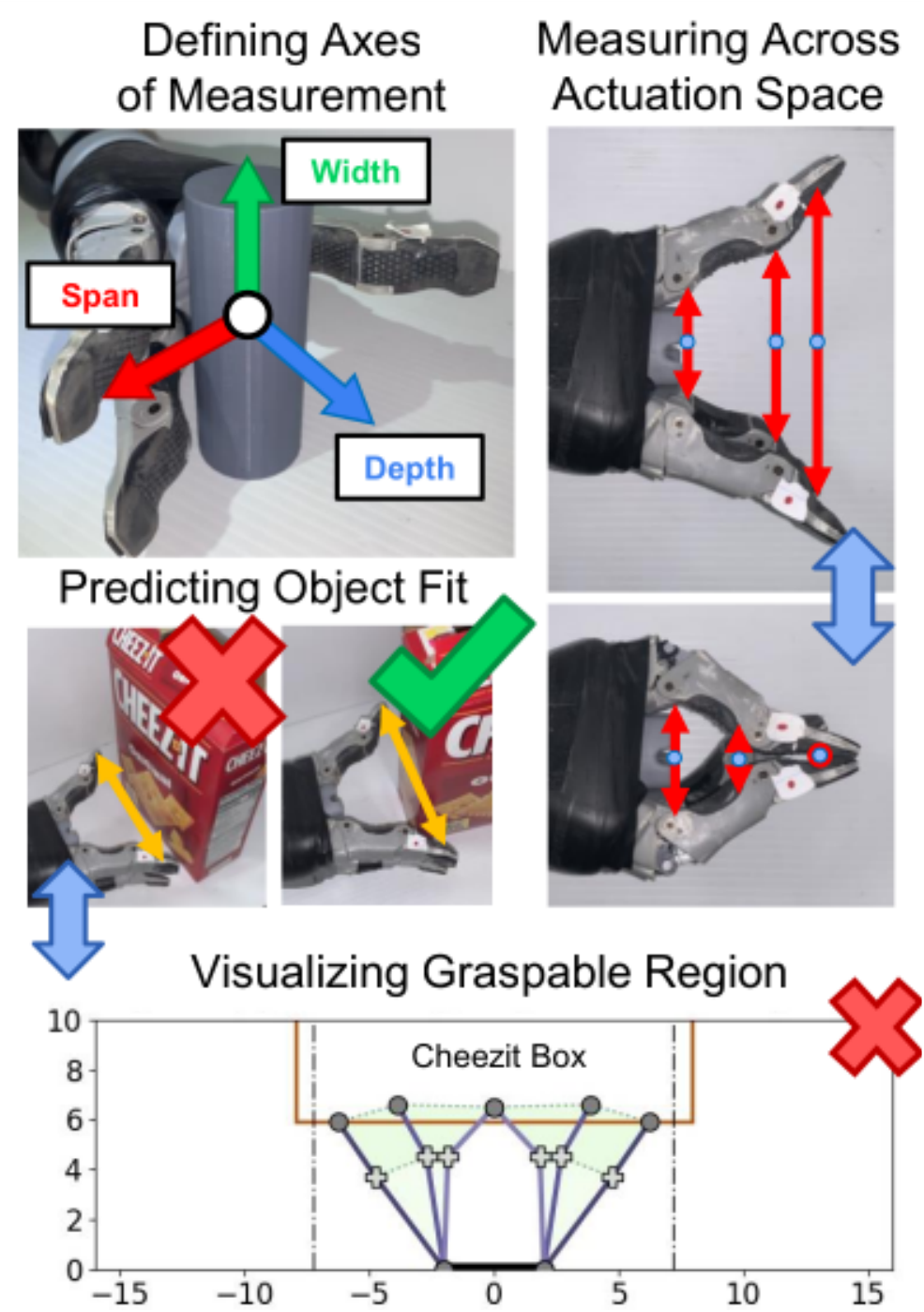}
    \caption{Defining and using the measurements. Top left: establishing the coordinate system using a functional grasp. Top right: taking measurements at different depths for the hand as it moves from open to closed. Middle Left: using the measurements to determine {\em a priori} if an object will fit in the hand. Bottom: linearly interpolating the measurements provides estimates of the graspable region.
    \vspace{-0.3cm}}
    \label{fig:intro}
\end{figure}

While it is conceptually simple to take a few measurements and linearly interpolate them, in practice the challenge is defining a scheme to do so. The approach should be both repeatable --- two people familiar with the hand should produce similar measurements, and generalizable --- it should be applicable to a wide variety of robot hand designs. Unfortunately, a robot hand's ``grasp size" is not just a function of finger link lengths and joint angles, but includes additional factors such as how movements of the fingers are coupled (under-actuated hands) and where and how the fingers make contact with the object. These are, in turn, determined by joint compliance and the hand actuation scheme~\cite{bib:grasp-and-contact}. Moreover, the {\em dimensions} of the space enclosed by the fingers and palm are coupled and depend on the type of grasp and how the hand is posed relative to the object. For example, a robot hand might be able to grab an American football lengthwise --- but not a soccer ball --- because depth from the finger tips to the palm is too shallow. Our primary contribution is a {\em functional}, object-centric approach for specifying these measurements that is simple and efficient in practice.

More specifically, we use an applied scenario --- grasping a cylinder or sphere resting on a table --- to define the measurement axes (see Fig.~\ref{fig:intro}). We then measure the hand's dimensions at two (or more) points in the grasp (opened to closed). The specifics of how to actuate the hand, where to measure the dimensions, and where to measure the contact points, are left to the human expert. To ensure repeatability we provide functional descriptions of the measurement points and contact surfaces and require documentation of them. The entire process is relatively efficient (less than an hour) and can be done with either a physical hand or a physical simulation.

We provide additional software tools to visually display the measurements, calculate if an object of a specific size is graspable, and define object sizes that are  small, medium, or large with respect to the hand. The latter is particularly useful when comparing hand functionality across two different hand morphologies. 

\textbf{Contributions:} 1) A measurement system which provides an intuitive, quantitative representation of a hand's graspable space and how it changes as the hand's fingers actuate, and 2) the ability to assess an object's relative size with respect to a hand's dimensions.

\begin{figure}
    \centering
    \includegraphics[width=0.98\linewidth]{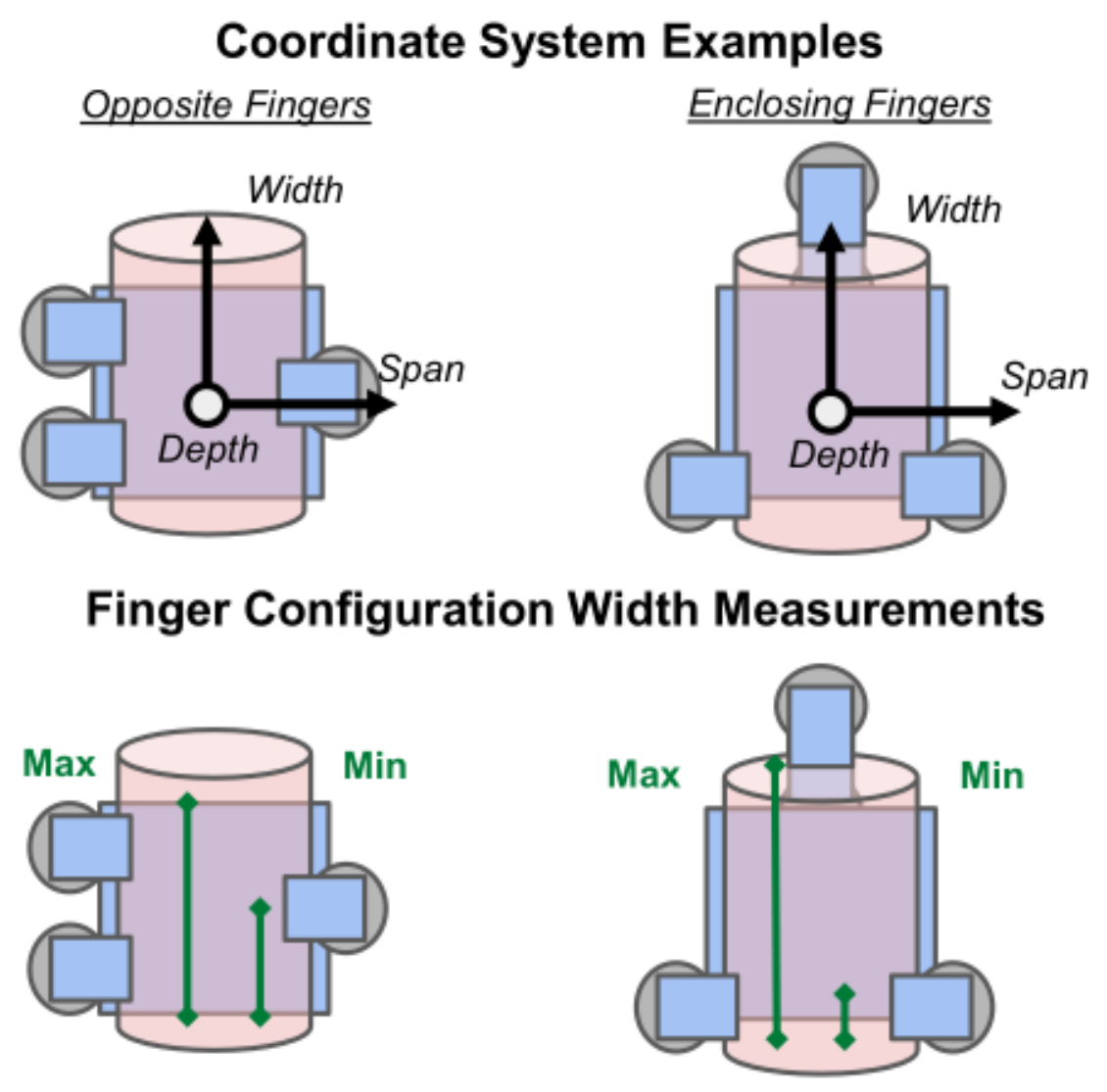}
    \caption{Our measurement standard captures how the fingers and palm enclose an area defined by three axes. These axes are defined by placing the hand in a position to grasp a hypothetical cylinder on a table. The \textit{span} axis measures {\em across} the palm, the \textit{depth} axis measures out of the palm, and the \textit{width} axis measures the cylinder's height. Shown are two common hand configurations for grasping the cylinder.
    \vspace{-0.3cm}}
    \label{fig:axes}
\end{figure}

\section{Related Works}
\begin{figure*}[ht]
    \centering
    \includegraphics[width=0.98\textwidth]{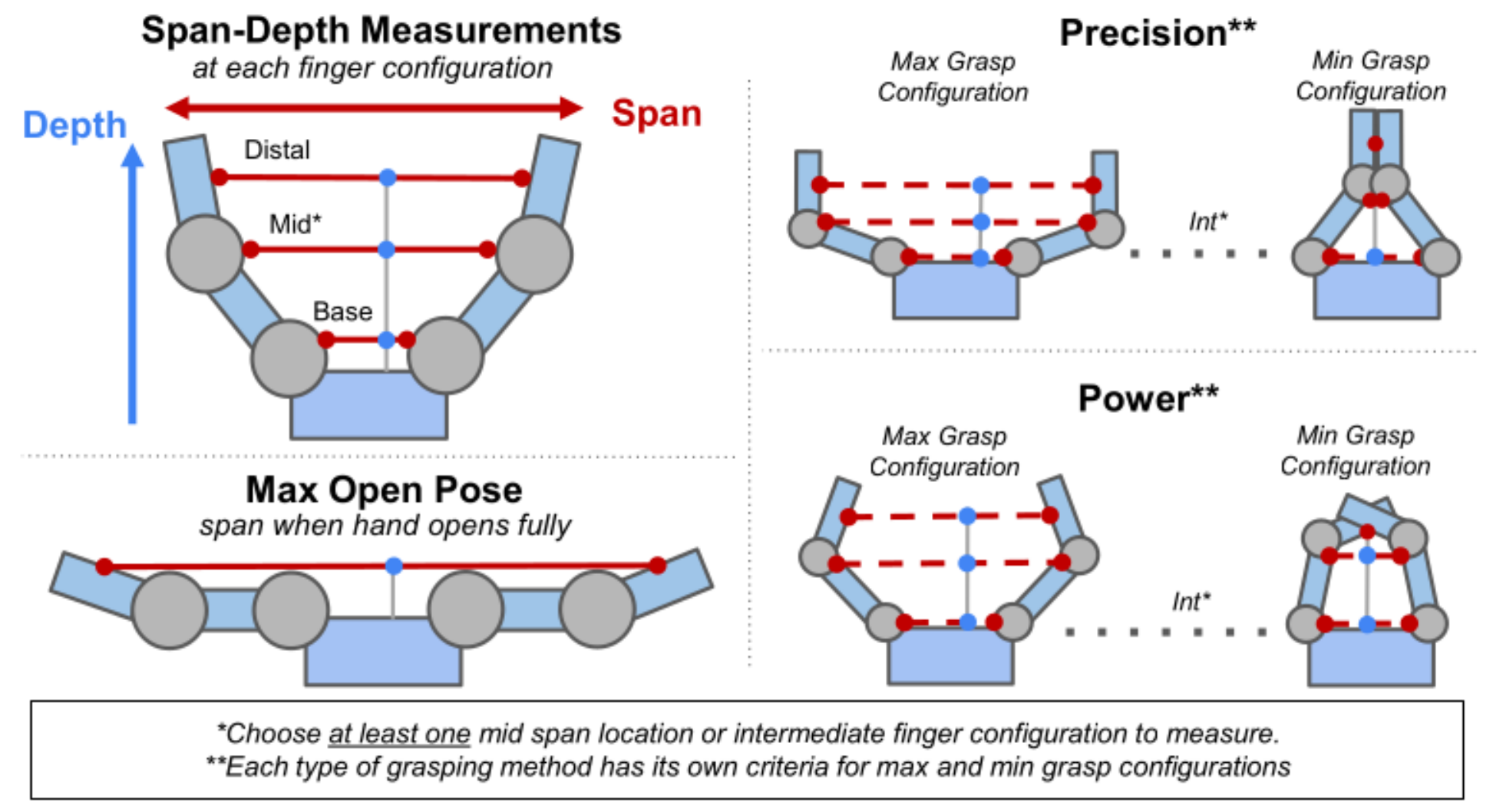}
    \caption{\textit{At least two measurements for at least two finger configurations.} |  \textit{Top Left:} We require sufficient span-depth measurements to represent the space defined by a finger configuration. \textit{Bottom Left:} Max Open pose is a one-time measurement taken at the fully open pose. \textit{Right:} We require sufficient finger configuration measurements, going from the largest functional grasp to the smallest functional grasp to capture how the space changes as the fingers actuate. Show are two types of grasps:  precision (top right) and power (bottom right). 
    \vspace{-0.3cm}}
    \label{fig:measurements}
\end{figure*}
\subsection{Hand Measurements in Industry}
Without access to physical robot hands, prospective buyers must first consult commercial product pages to assess if a hand will meet their needs. Product pages for robot hands vary widely in the information that they provide, focusing mainly on force output or payload weight information. 
We sample five product pages (Barrett~\cite{bib:barrett-specs}, Jaco2~\cite{bib:jaco2-specs}, Robotiq 2F-85~\cite{bib:robotiq2f-specs}, Robotiq 3F~\cite{bib:robotiq3f-specs}, and Yale Openhand~\cite{bib:yale-openhand} hands) to show a selection of representations for a hand's grasping limits. 

The Barrett hand provides link length and joint limit specifications in the advanced documentation (see ``kinematics, joint ranges, conversion factors'' in wiki at~\cite{bib:barrett-specs}). The Robotiq 2F-85 product page provides `stroke length', that is the range of distal link travel~\cite{bib:robotiq2f-specs}. The Robotiq 3F adds `object diameters for encompassing' in their specifications~\cite{bib:robotiq3f-specs}. For the Jaco2 and Openhand, prospective users can access CAD/STL models to get a sense of their size but not how they actuate~\cite{bib:jaco2-specs,bib:yale-openhand}.

Each product page provides information for inferring robot hand size; however most require expert knowledge to process that information (for example, kinematics and modelling). Our work is intended to supplement the detailed specifications with practical measurements that do not requiring expert knowledge to use.

\subsection{Hand Measurements in Academia}
In academia, the prevailing focus has been to define `object-ranges'~\cite{bib:guesteditorial} that a hand can grasp using object-sets (a  sample:~\cite{bib:ycb-objectset,bib:amazon-objectset, bib:grocery-objectset}). Our work differs by approximating the hand's space, using sets of measurements, across its actuation.

Most workspace characterizations focus on the dynamics of hand-object interactions within a hand's space, for example push-grasping~\cite{bib:push-grasping}, object resting  positions~\cite{bib:ocean-hand, bib:power-manip}, and others~\cite{bib:nist-benchmark, bib:human-precision-workspace, bib:learning-kinematic-workspace}. Our work complements these methods because it assesses whether an object can \textit{fit} inside a hand with contact.

Assessing whether an object fits inside a hand's space is different than predicting grasp success based on an object's size, shape, and position (such as~\cite{bib:ammar-object-sizes}). Instead, one could use our work for 1) understanding how well an object can \textit{fit} inside the hand's space, 2) understanding the relative size relationship between an object and a hand, and 3) assessing whether antipodal contact on the object is possible.

\section{Hand-measurement Benchmark}
\label{sec:methods}
Our goal is to provide an approximation of the space enclosed by the hand's palm and fingers. This requires defining a coordinate system; the challenge is that not all hands have a ``natural'' coordinate system because the fingers can enclose an object from arbitrary directions. Our solution is an object-centric, functional definition of the coordinate system. We place three axes of measurement (span, depth, and width) onto a cylinder resting on a table and then position a hand in the best configuration for grasping the cylinder from the side. This enables transfer of the axes to the hand~(see Fig.~\ref{fig:axes} and Section~\ref{sec:axes}).

Given a specific finger configuration, we approximate the space enclosed by the hand space using~(at least) two measurements at two different depths (see left of Fig.~\ref{fig:measurements}, Sec.~\ref{sec:span-depth}). This process captures the {\em boundary} of the shape created by finger actuation. We further require measuring at least two finger configurations which represent the largest and smallest (valid) grasp possible with the hand (see right of Fig.~\ref{fig:measurements}, Sec.~\ref{sec:configs}). This captures how the enclosed space changes as the fingers close. 

The space enclosed by a hand can change based on the type of grasp. Moreover, the fingers need to be placed so that they {\em can} grasp an object. We propose simple, yet functional guidelines for choosing finger configurations to avoid extensively testing each hand. We specify two types of grasps: precision (Sec.~\ref{sec:precision}) and power (Sec.~\ref{sec:power}), which primarily differ on whether or not the fingers contact at the mid-line of the cylinder or enclose it. We propose separate finger configuration guidelines for each grasp type (see Fig \ref{fig:power}). 


We provide software tools to 1) calculate intermediate grasp regions using linear interpolation, 2) to visualize those regions, 3) determine if an object will fit (and if so, how well) within the hand, and 4) qualitative  measurements of the relative size of an object with respect to the hand's grasp space (Sec,~\ref{sec:rel-size}). 

These functional definitions and guidelines represent a compromise between repeatability (two different measurers will produce the same measurements) and generalizability (applicable to a wide variety of hand designs). We rely on the measurer (either a roboticist or the manufacturer) to determine the hand configurations that best match the intention of the grasp. To balance accuracy with measurement time, we require sufficient measurements so that the linear approximation is accurate within a given percentage. To increase repeatability we provide additional documentation guidelines designed to clarify those choices (Sec.~\ref{sec:accountable}).

Detailed instructions, supporting software, and example measurements are provided on github~\cite{bib:github}.

\subsection{Functional Axes: Span, Depth, and Width}
\label{sec:axes}
To define the coordinate system we use for measuring the hand we first attach the coordinate system to a cylinder (or sphere) resting on a table (see Fig.~\ref{fig:axes}). The measurer then places the hand and configures the fingers in order to grasp the cylinder from the side. The intention is to define a clear in-out from the palm (depth) and left-right finger closing direction (span). While no physical object is required it can be helpful to use one in order to set up this positioning. The functional definition of the axes is as follows:

\begin{figure}[t]
    \centering
    \includegraphics[width=0.49\textwidth]{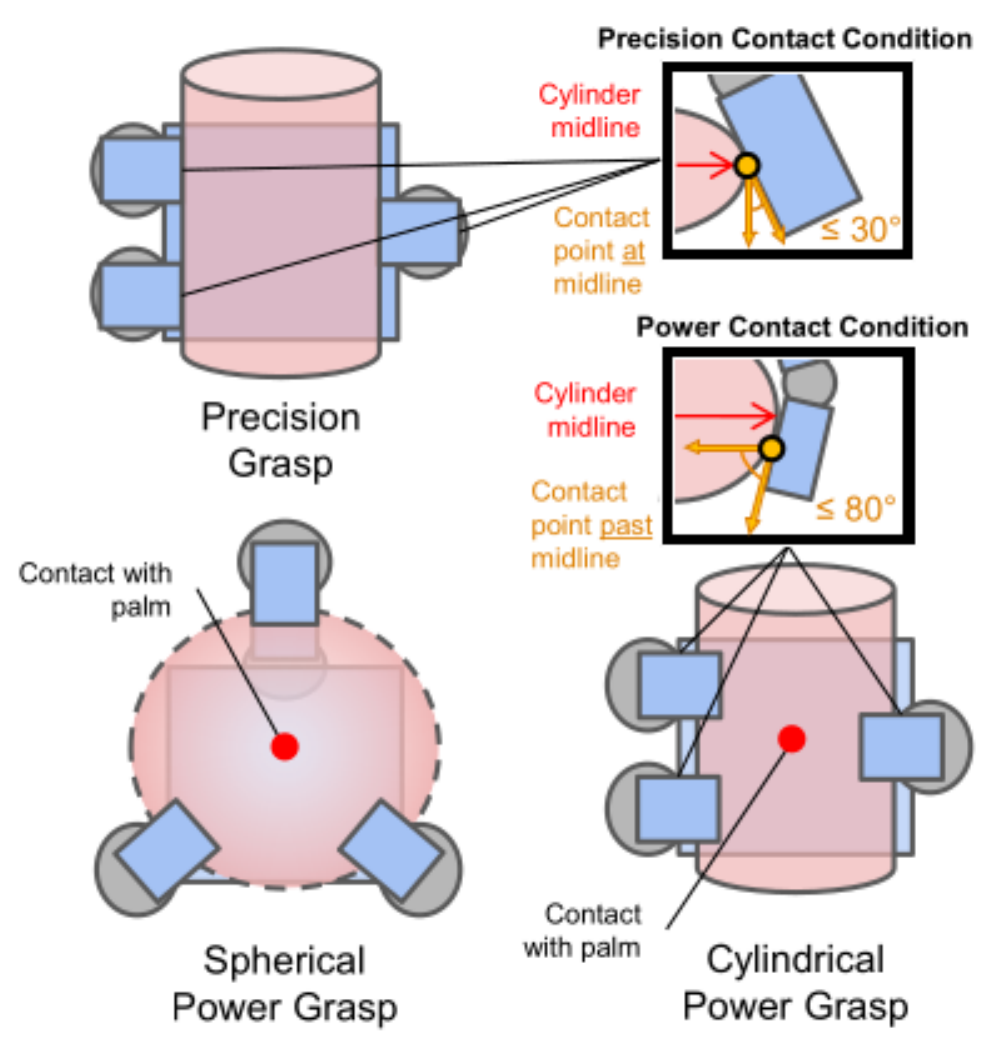}
    \caption{Contact configurations for a precision (top) and power (bottom) grasp. The power grasp can be performed with either a cylinder (right) or sphere (left). 
    \vspace{-0.7cm}}
    \label{fig:power}
\end{figure}

\noindent {\bf Span }--- is the axis parallel to both the reference cylinder's top/bottom faces and the palm's normal --- i.e., a line passing through the middle of the cylinder parallel to the table. Moving in the span axis is akin to moving the reference cylinder between the fingers across the table surface. This axis measures the largest object the fingers can span; the primary open/close motion of the fingers should be in this axis.

\noindent {\bf Depth }--- is the axis orthogonal to the plane of the palm. Moving in the depth axis is akin to moving the reference cylinder closer to, or farther from, the palm. Span and Depth are typically coupled under finger actuation; ``opening'' the fingers in the span direction will usually change the depth. 

\noindent {\bf Width }--- is the axis normal to the table top. Moving in the width axis is akin to lifting the cylinder straight up off the table. The width axis measures the height of objects that can be grasped by a hand. For most hands, the width measurements remain constant throughout actuation.

\subsection{One-time Measurements}
\label{sec:onetime}
There are three measurements that are taken once and do not change with finger actuation. These are: the Maximum Open, and the Minimum and Maximum Width measurements. The Maximum Open measurement is used
to record the span when the fingers are completely open.
The Minimum Width measures the shortest object the hand could pick up off of the table. The Maximum Width represents the tallest object or width of the hand, depending on finger layout.

The Maximum Open measurement measures the distance between the ends of the distal links, without taking grasping into account. For this measurement, we open the hand as far as it will go. This is the finger configuration typically used in other work~\cite{bib:push-grasping, bib:ocean-hand}. 

It is important to consider the height an object has to be for the hand to have antipodal contact
(see Fig.~\ref{fig:ycbs}). Minimum Width is the distance from the tabletop to the lowest contact point at which fingers on opposite sides would connect with the object. 

Our definition for Maximum Width depends on the hand orientation. In general, Maximum Width represents the tallest object which could fit inside the hand (i.e. Fig. \ref{fig:power}, the tallest object that fits under the top finger). 
For hands without an upper limit (such as left of Fig. \ref{fig:power}), the tallest object could (theoretically) be infinite. Instead, in these cases, we record the hand's palm width with a `+' after the measurement.

\begin{figure*}
    \centering
    \includegraphics[width=0.99\textwidth]{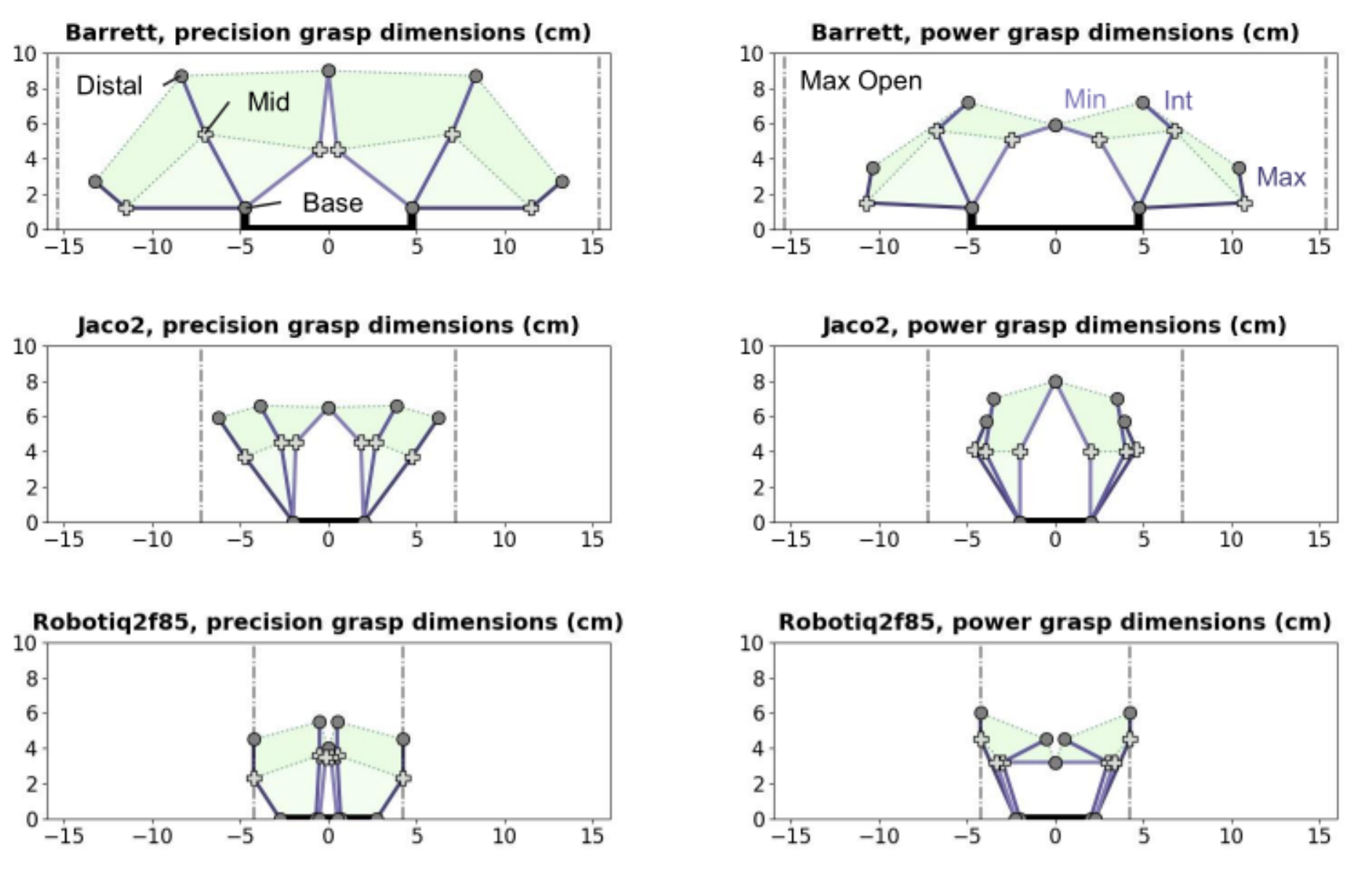}
    \caption{Precision (left) and Power (right) hand measurements for the Barrett (top), Kinova Jaco 2 (mid), and Robotiq 2f-85 (bottom) hands. Plusses are used for the mid span-depth pair to reinforce that they do not \textit{have} to represent a measurement at a joint. 
    \vspace{-0.3cm}}
    \label{fig:dims_results}
\end{figure*}

\subsection{Span-Depth Measurements at a given Finger Configuration}
\label{sec:span-depth}

When measuring a hand configuration we require at least two span-depth measurements (Base and Distal) in order to approximate the enclosed grasp shape (see left of Fig.~\ref{fig:measurements}). The locations of these measurements are based on depth. 
Each measurement records the Depth and the Span, for a minimum of two  measurement pairs.

The Base measurement is taken at the center of the proximal joints. If the proximal joints are flush with the palm, the base measurement is taken at a depth of zero (at the palm). 

The Distal measurement occurs at the distal link. We provide two locations: at the tip or at the center of the distal link (on the fingerpad surface), depending on the finger configuration used to accomplish the grasp. The tip should be used only when the tip of the finger is in contact with the hypothetical cylinder (in this case the fingerpad surface is perpendicular to the contact). 

The Mid measurement(s) (there can be more than one) are taken between the Base and Distal measurements. They are intended to capture how the enclosed space widens between the Base and Distal measurements. The measurer should choose the number and placement of these measurements to capture maxima (and possibly minima) of the changes in Span with respect to Depth in the hand's space. For the hands measured in this paper we chose one Mid measurement and placed it at the distal joint as an example.

\subsection{Finger Configurations to Measure}
\label{sec:configs}
We approximate how the enclosed space changes as the hand is actuated using at least two finger configurations. The finger configurations correspond to the Maximum and Minimum {\emph functional} grasps (see right of Fig.~\ref{fig:measurements}).
These functional configurations must represent grasps, and therefore are not necessarily the widest/narrowest figure configurations.

Intermediate finger configurations represent any finger configuration between the Max and Min configurations, but they are not required. The intent of the Intermediate configurations is to approximate how the Span-Depth measurements change as the fingers are actuated. Using linear interpolation we can generate approximate Span-Depth measurements at any point in the actuation. Where this approximation differs too much from the actual measurements, an intermediate measurement should be added. For the hands measured in this paper we chose one intermediate measurement at the  half-way actuation point as a demonstration.


\begin{figure*}
    \centering
    \includegraphics[width=0.98\linewidth]{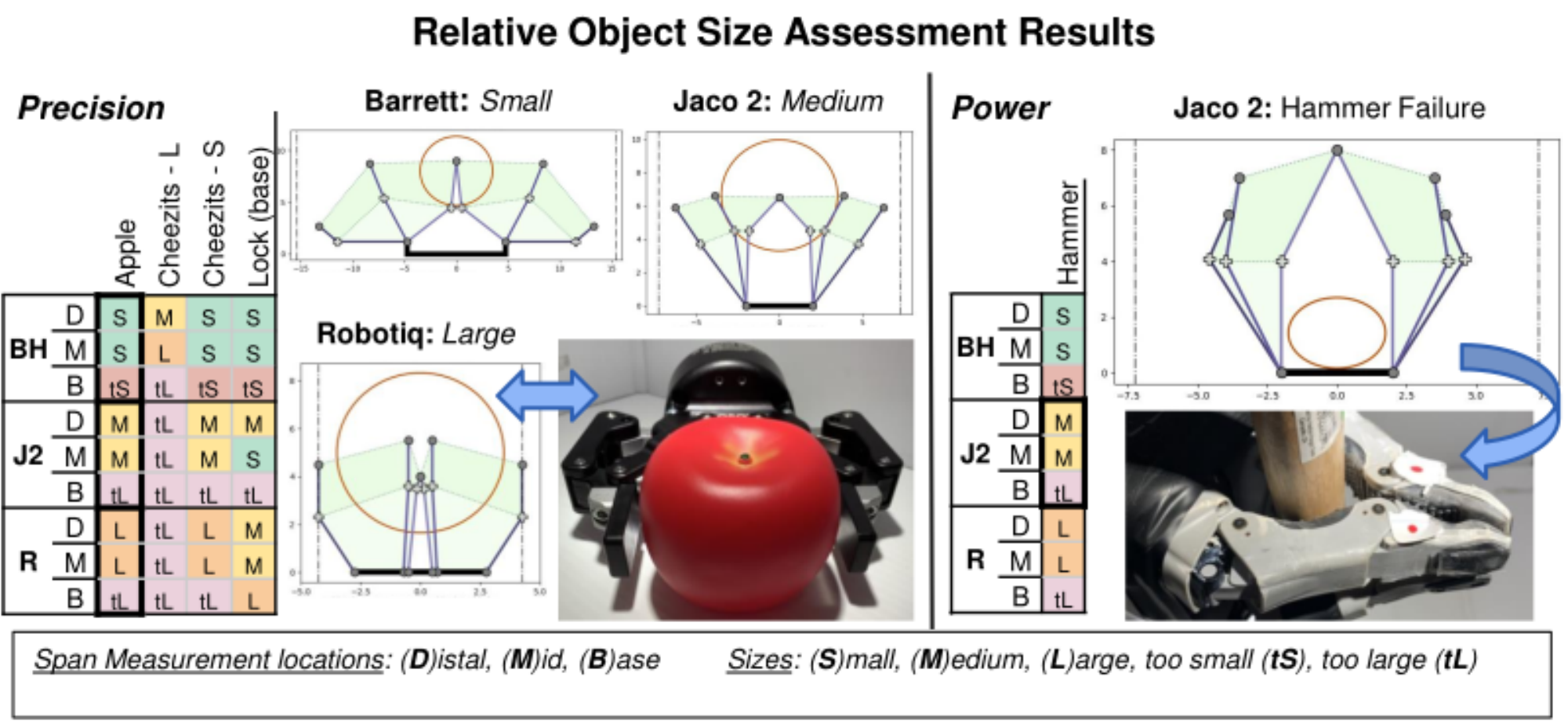}
    \caption{Relative object size assessments for each hand and object. Bold outlined assessments are shown visually. The Large Assessment is shown on the Robotiq 2F-85, and the failure of the kinova Jaco 2 hand to power grasp the hammer shaft are also shown.
    \vspace{-0.3cm}}
    \label{fig:rel_obj_assessment}
\end{figure*}

\subsection{Precision Grasp Measurements}
\label{sec:precision}

We define the precision grasp functionally as a stable grasp with finger contacts at the centerline of an ovoid shape held in the distal links. The distal link contact surfaces should be oriented at most 30\degree\ outward relative to the depth axes (see Fig.~\ref{fig:power}) as a functional guideline. The intent here is to eliminate grasps in which the fingers contact the object without sufficient frictional force to keep the object from ``popping'' out of the grasp.

\subsection{Power Grasp Measurements}
\label{sec:power}
We define the power grasp functionality as a stable grasp with at least three contact points on an ovoid object. This is typically one contact point on the palm and the others at the fingers. The finger contact points should be {\em past} the centerline of the object and at an angle of at least 80\degree\ relative to the distal measurement line (see middle right of Fig.~\ref{fig:power}) as a guideline.

We define two power grasp variations: A spherical power grasp that fully encloses the object from three (or more) directions, and a cylindrical power grasp that encloses a (2D) circular area. 
Hands may be capable of none, one, or both grasps --- which variations to measure is determined by the measurer. 

The Span-Depth measurements are slightly different for the spherical power grasp because the contact points are no longer planar. Instead of a single pan length measurement we measure the area of a disk that fits at each depth, effectively merging span and width.


\subsection{Improving Measurement Repeatability}
\label{sec:accountable}
We include here brief recommendations on how to make the measurements repeatable. 

First, for physical hands it may be difficult in practice to hold the fingers in the desired configurations, particularly for underactuated hands performing power grasps. In this case we suggest using an actual object to keep the hand in position. 

Second, if a kinematic, actuated CAD model exists then one can alternatively measure a hand virtually.

Finally, we recommend taking a picture, looking down on the table, of each finger configuration with a grid or ruler showing where the measurements (span and depth) were taken. For spherical measurements this may require a second picture taken looking at the palm.

\subsection{Software Tools}
\label{sec:rel-size}

We provide additional software tools to 1) interpolate the measurements, 2) determine if an object will fit in the hand, and 3) determine qualitative (small, medium, large) object-hand region definitions. All of these are parameterized by the grasp type $t$.

Span $s$ (optionally width $w$ for spherical grasp types) is defined as a piece-wise linear function of the depth, $d$. Each hand configuration defines $n \geq 2$ sets of depth-span(width) measurements. These sets are, in turn, linearly interpolated based on the percentage actuation, $a$:

\begin{eqnarray}
(s_a,d_a)_n & = & \mbox{ConfigInterp}(t, a) \\
s & = & \mbox{SpanInterp}(t, (s_a,d_a)_n, d)
\end{eqnarray}

\noindent Given an object's cross-sectional measurements ($o_s, o_d$) and optional height ($o_w$), and desired grasp type $t$, we define a simple search function that determines an ideal actuation percentage for enclosing the object. This function returns the size of the object with respect to each axes $x$ as well as the best actuation percentage $a$ and depth $d$ to place the object's center at:
\begin{eqnarray}
    (tS/S/M/L/tL)_x, a, d & = & \mbox{Fit}(t, o_s, o_d, o_w) 
    \label{eq:fit}
\end{eqnarray}

\noindent The size measurements are calculated based on the maximum $M_x$ and minimum $m_x$ values for each axes $x$: 
\begin{eqnarray}
s_x & = & \frac{O_x - m_x}{M_x - m_x} 
\end{eqnarray}

\noindent
We define the size ranges as Too Small $(s_x < 0)$, Small $(0 \leq s_x < 0.3)$, Medium $(0.3 \leq s_x < 0.7)$, Large $(0.7 < s_x \leq 1.0)$, and Too Large $(s_x > 1)$. The closer $s_x$ is to 0 or 1, the less tolerance there is for error when grasping the object.

Inverting the mid-point of these ranges (eg, using $s_x = 0.15$ and calculating $O_x$) is a straightforward way to define a canonical ``small'' object size with respect to the hand.

Fig.~\ref{fig:dims_results} illustrates our approach for plotting these measurements in a spatially useful way.

\section{Demonstrating and using  Measurements}
\label{sec:results}
\begin{figure}
    \centering
    \includegraphics[width=0.99\linewidth]{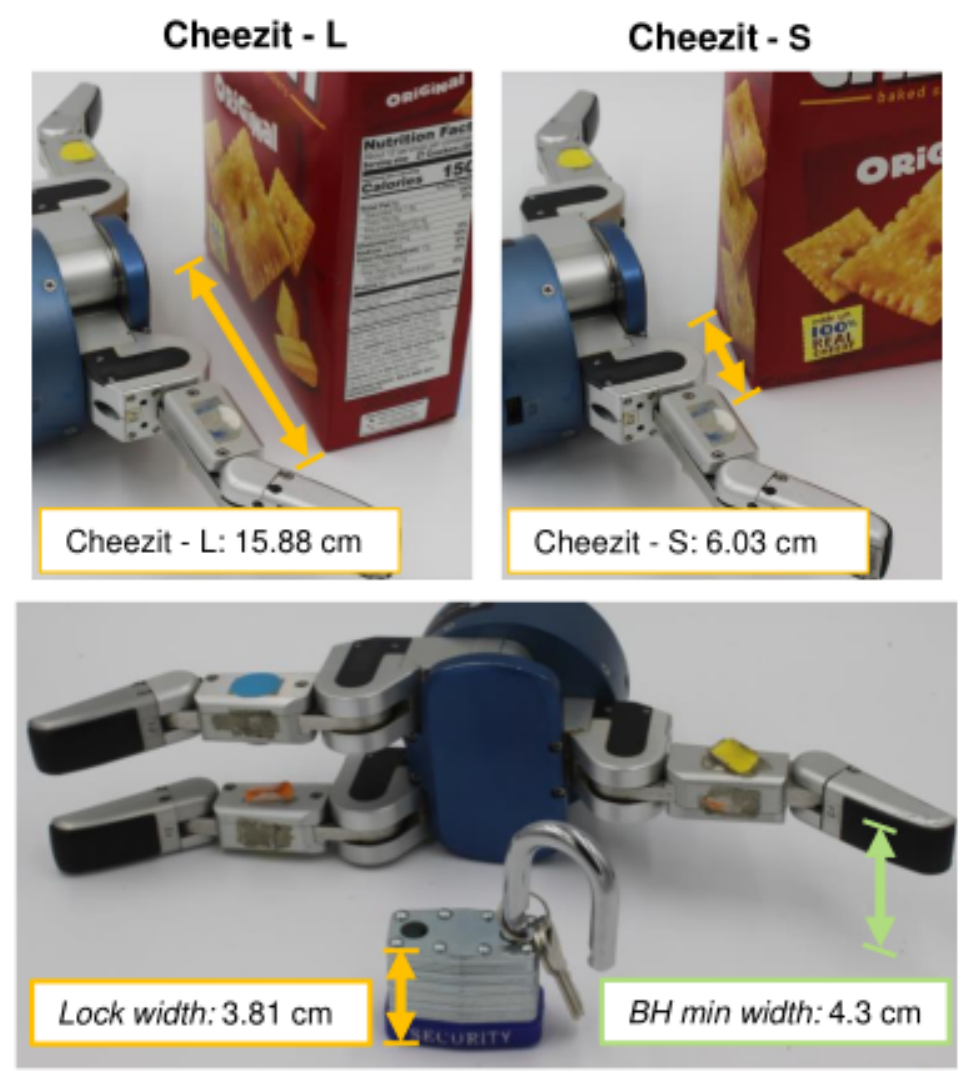}
    \caption{\textit{Top}: Grasping configurations for two orientations of the box, with sizes. \textit{Bottom}: The base of lock is actually too low to be grasped by the Barrett hand.
    \vspace{-0.3cm}}
    \label{fig:ycbs}
\end{figure}

We demonstrate our measurement system with a hand-buying scenario. In this example, we evaluate how well three commercial robot hands might work for grasping specific objects. Our three hands are: a modified Barrett (puppet version from \cite{bib:john-phig}), the Kinova Jaco2, and the Robotiq 2f-85 hands. 
Our example tasks are as follows: 1) grasping an apple and Cheez-It box, 2) grasping a lock with the intent to lock it on a chest, and 3) grasping a hammer by its handle for use (see Fig.~\ref{fig:ycbs}). All objects in this example are from the YCB object set. 

For each object we stipulated a specific grasp type in order to accomplish a given task: (i)~a precision grasp for the apple and the box, (ii)~grasping the base of the lock so that it can thread the bar through the chest hole, and (iii)~a power grasp around the hammer handle.

First, we measure the hands and the objects, using one Mid measurement and one Int configuration as an example. The resulting plots are shown to scale in Fig.~\ref{fig:dims_results} for both precision and power grasps. These plots clearly show the overall size differences between the three hands, as well as how the depth and span ratios change with actuation. Note that the shaded regions indicate  areas where link contact \textit{can} occur on an object. 

Fig.~\ref{fig:rel_obj_assessment} summarizes the relative object size assessment for the apple and hammer with the desired grasp type.

\subsection{Assessment Results}
We first evaluate how well the apple fits into each hand. From our size assessment (Eq.~\ref{eq:fit}) the apple's relative size is small (Robotiq), medium (Jaco2), and large (Barrett). We provide hand-dimension plots for each hand with the apple's cross-section overlaid in the top of Fig.~\ref{fig:rel_obj_assessment}. Note how little room is left in the Robotiq's hand space with the apple inside it. This implies that the hand may have difficulty grasping the apple in the presence of noise. In contrast, the Barrett hand's large maximum span relative to the apple's size might put it at a disadvantage grasping in cluttered environments.

We evaluated both possible grasp directions of the Cheez-It box to demonstrate how relative size assessment could also be used in the context of grasp planning. We found that the long side is too large for the Jaco2 and Robotiq hands. For the Barrett hand, a medium assessment at the distal location suggests that the Barrett hand can comfortably grasp the box at the distal links. All hands could grasp the box on its short side, suggesting this to be the preferred side to grasp the object.

The lock fits inside each hand when considering measuring span and depth. However, the top of the lock's base, when placed on the table, is shorter than the minimum width of the Barrett hand (see Fig.~\ref{fig:ycbs}), meaning it can't be picked up. 

The hammer, despite appearances, is actually too small for the jaco 2 hand (see top-right of Fig.~\ref{fig:rel_obj_assessment}).


\subsection{Benchmark Limitations}

One limitation of our current approach is that we do not support ``negative'' distal span configurations, in which the fingers cross over each other. An example of this is the Barrett Hand, which is able to power grasp the hammer by letting the fingers wrap past each other. We could, theoretically, allow this configuration (and negative measurements) but it would complicate both the measurement process and make the visualizations more difficult to understand.

Another limitation is that this measurement system does not handle hands with complex actuation patterns~(such as hands using synergies, e.g.~\cite{bib:astrobee-hand}) well. Our challenge with the limitations is to find a way to practically measure the hand and intuitively represent that space. 
Future work can explore these two limitations.

\subsection{Advantages of Normalizing hand design}

Normalizing object sizes can provide the basis for a more direct comparison of hand designs. Our process extends the current practice, which is to normalize objects by finger length~\cite{bib:power-manip}, to instead normalize the object to the \textit{whole} hand space. Additionally, the normalization of object-hand relationships can also impact human grasping research for haptics and hand rehabilitation because it provides a method for optimizing control/object size based on a human subject's hand size~\cite{bib:human-grasp-object-properties}.


\section{Conclusion}
In this work we introduced a hand measurement standard which approximates the area of a hand's space as the hand's fingers actuate in free space.
We visualize these measurements in a user-friendly manner for assessing object sizes in a hand's space. 
Our process also provides a method of normalizing hand-object size relationships to compare between hand designs.
In future work, we plan to continue to refine this standard for hand designs with more complex span-depth-width ratios, and to investigate the implications that object size has on grasp success and manipulation ability.


\section*{Acknowledgment}
This work supported in part by NSF grants CCRI 1925715 and RI 1911050. We also thank Adam Norton and Brian Flynn at the University of Massachusetts Lowell for profitable discussions and for critiquing earlier iterations of our measurement standard.

\ifCLASSOPTIONcaptionsoff
  \newpage
\fi

\bibliographystyle{IEEEtran}
\bibliography{main}

\end{document}